\documentclass{article} 
\usepackage{iclr2023_conference,times}

\usepackage{amsmath,amsfonts,bm}

\def\eqref#1{equation~\ref{#1}}

\def\1{\bm{1}}

\def\ry{{\textnormal{y}}}

\def\rvc{{\mathbf{c}}}

\def\rvx{{\mathbf{x}}}
\def\rvy{{\mathbf{y}}}

\DeclareMathAlphabet{\mathsfit}{\encodingdefault}{\sfdefault}{m}{sl}
\SetMathAlphabet{\mathsfit}{bold}{\encodingdefault}{\sfdefault}{bx}{n}

\DeclareMathOperator*{\argmin}{arg\,min}

\usepackage{url}

\usepackage[utf8]{inputenc} 
\usepackage[T1]{fontenc}    
\usepackage{url}            
\usepackage{booktabs}       
\usepackage{amsfonts}       
\usepackage{nicefrac}       
\usepackage{microtype}      
\usepackage{xcolor}         
\usepackage{microtype}
\usepackage[pdftex]{graphicx}
\usepackage{subfigure}
\usepackage{wrapfig}

\usepackage{amsmath}
\usepackage{amssymb}
\usepackage{mathtools}
\usepackage{amsthm}
\usepackage{adjustbox}
\usepackage{fixltx2e}
\usepackage{setspace}
\usepackage{booktabs}
\usepackage{array}
\usepackage{tabularx}
\usepackage{calc}
\usepackage{makecell}
\usepackage{bbm}

\newcommand{\lm}{p_{\textrm{LM}}}

\newcommand{\cP}{\mathcal{P}}
\newcommand{\cA}{\mathcal{A}}
\newcommand{\dmeta}{\mathcal{D}_{\textrm{meta}}}
\newcommand{\indep}{\perp \!\!\! \perp}

\newcommand{\cD}{{\mathcal{D}}}
\def\rvx{{\mathbf{x}}}
\def\rvy{{\mathbf{y}}}

\def\rvc{{\mathbf{c}}}

\def\ry{{\textnormal{y}}}

\newcommand{\cS}{{\mathcal{S}}}
\newcommand{\bbR}{{\mathbb{R}}}

\definecolor{link-color}{RGB}{34, 180, 84} 
\definecolor{cite-color}{RGB}{46, 134, 190}
\usepackage[colorlinks=true,linkcolor=link-color,citecolor=cite-color]{hyperref}

\usepackage[textsize=tiny]{todonotes}

\usepackage{color, amsfonts, comment}

\title{LMPriors: Pre-Trained Language Models as
Task-Specific Priors}

\author{%
  Kristy Choi\thanks{Equal contribution}, Chris Cundy$^*$, Sanjari Srivastava, Stefano Ermon \\
  Department of Computer Science\\
  Stanford University\\
  \texttt{\{kechoi, cundy, sanjari4, ermon\}@cs.stanford.edu} \\
}

\iclrfinalcopy 
\begin{document}

\maketitle

\maketitle

\begin{abstract}
Particularly in low-data regimes, an outstanding challenge in machine learning is developing principled techniques for augmenting our models with suitable priors. This is to encourage them to learn in ways that are compatible with our understanding of the world. But in contrast to generic priors such as shrinkage or sparsity, we draw inspiration from the recent successes of large-scale language models (LMs) to construct \emph{task-specific priors} distilled from the rich knowledge of LMs. 
Our method, Language Model Priors (LMPriors), incorporates auxiliary natural language metadata about the task---such as variable names and descriptions---to encourage downstream model outputs to be consistent with the LM's common-sense reasoning based on the metadata.
Empirically, we demonstrate that LMPriors improve model performance in settings where such natural language descriptions are available, and perform well on several tasks that benefit from such prior knowledge, such as feature selection, causal inference, and safe reinforcement learning. 
\end{abstract}

\section{Introduction}
\label{intro}
\begin{wrapfigure}{r}{0.5\textwidth}
\vspace{-0.7cm}
  \begin{center}
    \includegraphics[width=0.48\textwidth]{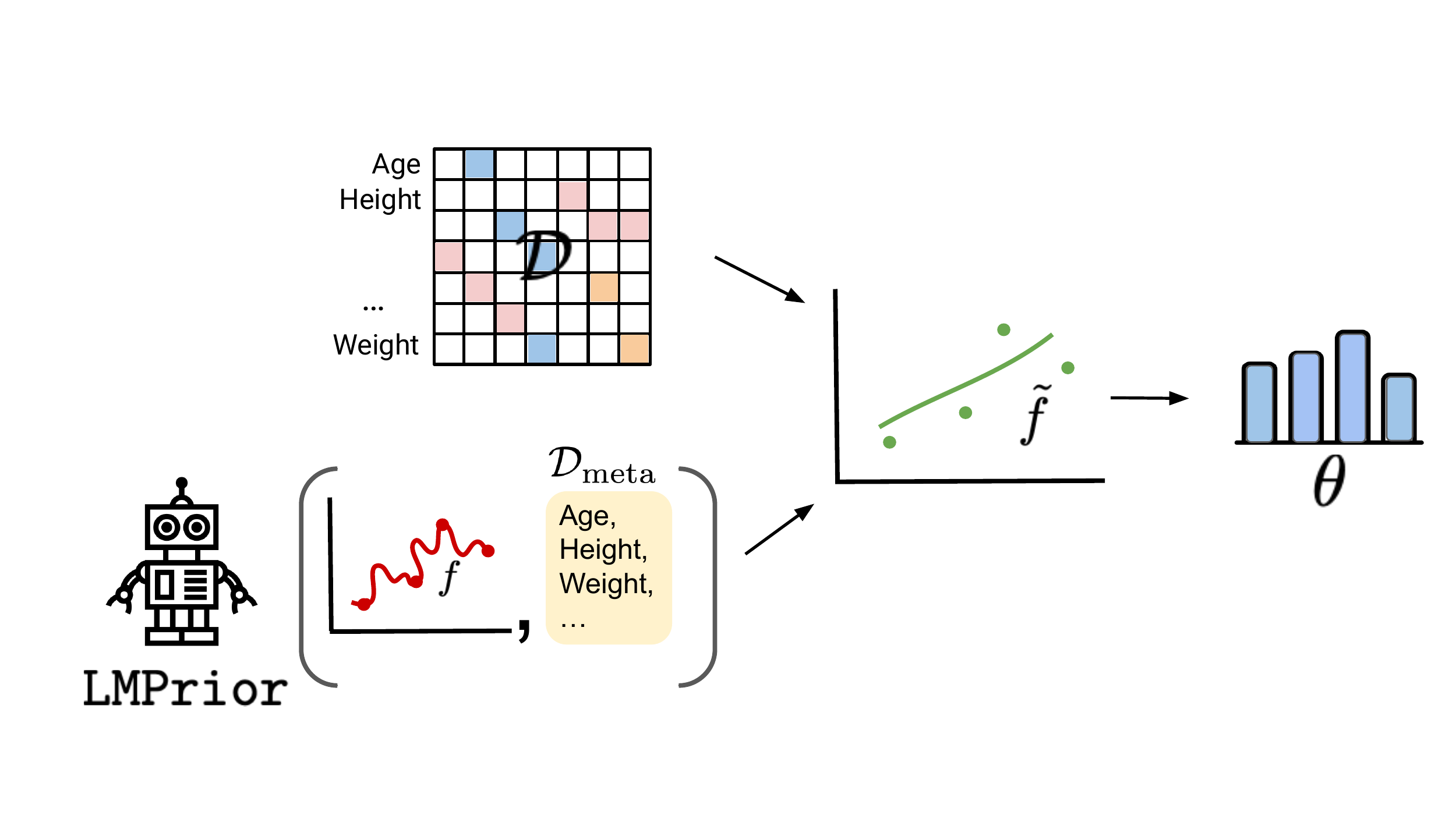}
    \label{fig:flowchart}
    \caption{A flowchart of the Language Model Prior (LMPrior) framework. We leverage the rich knowledge base of a pretrained LM to incorporate task-relevant prior knowledge into our learning algorithm $f$. Our method uses natural language metadata $\dmeta$ to return a specialized learner $\tilde{f}$, whose outputs given the dataset $\cD$ are encouraged to remain consistent with both the metadata and real-world knowledge as distilled in the LM.} 
  \end{center}
 \vspace{-0.5cm}
\end{wrapfigure}
Much of modern-day machine learning is \textit{data-driven}---given training examples, we aim to learn a function that minimizes an objective corresponding to a particular downstream task.
This paradigm has led to tremendous success in data-rich domains such as protein structure prediction for drug discovery \citep{jumper2021highly}, game playing \citep{mnih2013playing,silver2017mastering,bakhtin2021no}, automating medical diagnoses \citep{rajpurkar2017chexnet}, computational sustainability \citep{jean2016combining,attia2020closed}, and climate modeling \citep{sonderby2020metnet,ravuri2021skillful}.
However, the recent failures of such algorithms as in shortcut learning and vulnerability to adversarial examples
\citep{goodfellow2014explaining,guo2017calibration,xie2020adversarial,geirhos2020shortcut,d2020underspecification} seem to suggest that purely data-driven approaches have a long way to go from becoming truly \textit{intelligent} agents. 

One facet of intelligence which separates human agents from artificial ones is 
\textit{prior knowledge about the world} that can be combined with inferences derived purely from data.
Consider a prediction setting that aims to determine the length of one's commute time.
Although an algorithm may discover a relationship between commute time and favorite color, our intuition tells us that this relationship is most likely spurious.
Additionally, an autonomous driving agent may require several expert demonstrations before it learns that it should not veer off a cliff.
These failure modes are surprising to us precisely because they violate deeply-ingrained prior beliefs about how the world works. 
Artificial agents,
on the other hand, lack such grounding in real-world contexts and are thus limited in their ability to reason about the semantic relationships between entities present in data.
This problem becomes even more pertinent in low-data regimes where our algorithms are prone to overfitting.

\looseness=-1
Two key observations guide this work. 
The first is that auxiliary metadata, often in the form of natural language descriptions such as variable names that ground the features in real-world entities,
are becoming increasingly more abundant \citep{gebru2021datasheets}. 
The second is that in spite of this, most conventional learning algorithms are designed to \emph{ignore} this valuable information. This is understandable due to the subjective nature of prior information elicited from experts or algorithm designers, combined with the difficulty of scaling up approaches to thousands or millions of variables.
Inspired by the recent successes of large-scale pretrained language models (LMs) across a wide range of domains and data modalities \citep{radford2018improving,brown2020language,chen2021decision,lu2021pretrained,radford2021learning}, we propose to leverage the LM's rich knowledge base as a heuristic for prior knowledge about the world. This provides a pathway for algorithmically, scalably, and repeatedly generating relevant inductive biases from task-specific metadata.
Our framework, which we call Language Model Priors (LMPriors), then serves as a way to construct \emph{task-specific priors} tailored to any learning setting where natural language descriptions of the task are available.
We provide an illustrative flowchart of how LMPriors fit into the conventional machine learning pipeline in Figure~\ref{fig:flowchart}.

Empirically, we demonstrate that our LMPriors are able to perform well on a variety of downstream tasks which benefit from auxiliary sources of information. 
Concretely, our contributions are:
\begin{enumerate}
\item We introduce LMPriors, a framework for algorithmically incorporating semantically-relevant prior knowledge into learning problems via use of a prior distribution extracted from a LM.
\item We explicitly specify LMPriors for feature selection, causal discovery, and reinforcement learning tasks. Each LMPrior is a mapping from a set of task-specific metadata \({\cal D}_\text{meta}\) to a learning procedure with a bespoke inductive bias.
    \item We show empirically that LMPriors achieve strong performance on feature selection, causal discovery, and safe reinforcement learning tasks, and demonstrate that it can also serve as a useful preprocessing wrapper around existing algorithms to boost their performance.
\end{enumerate}
\section{Preliminaries}
\label{prelims}

\subsection{Neural Language Modeling}
\looseness=-1
Language modeling seeks to learn a probability distribution $\lm(\rvx)$ over variable-length sequences of text $\rvx =(\rvx_1, \ldots, \rvx_{\vert \rvx \vert})$, drawn from an underlying distribution $p_\textrm{text}(\rvx)$, such that $\lm(\rvx) \approx p_\textrm{text}(\rvx)$. 
Although several approaches exist for parameterizing $\lm(\rvx)$, conventional neural LMs posit an autoregressive factorization over $\lm(\rvx) = \prod_{i=1}^{\vert \rvx \vert} \lm(\rvx_i \vert \rvx_{<i})$ and are trained via maximum likelihood \citep{sutskever2014sequence,cho2014learning}.
When predicting the next token \(\rvx_i\), the preceding tokens \(\rvx_{<i}\) are known as the \emph{context} or \emph{prompt} $\rvc$.
Modern LMs are trained on large corpora consisting of billions of tokens over diverse sources of text including encyclopedias, news websites, emails, books, and scientific papers \cite{gaoPile800GBDataset2020}. In order to successfully predict the next token over such a diverse set of contexts, LMs implicitly possess rich knowledge about concepts in the training data. 
This allows them to solve a startling variety of tasks from simple descriptions of the task itself, a setting known as zero-shot learning \citep{brown2020language}. 
We seek to leverage this rich knowledge base as the foundation of our approach.

\looseness=-1
\textbf{Prompt design.}
Since the largest LMs are currently proprietary\footnote{We note that this status quo is quickly changing with open-source tools such as HuggingFace \citep{wolf2019huggingface}.}, we assume black-box access to the underlying LM and avoid cases where our method would need to fine-tune or access internal statistics (such as gradients or embeddings) of the model. Given this assumption, our control over the model's predictions relies entirely on our choice of prompt. Effective prompt design is a key challenge when utilizing modern LMs, and one that has been widely studied \citep{lester2021power,li2021prefix,qin2021learning,liu2021makes,zhao2021calibrate}.

\subsection{Task-Specific Knowledge in Data-Driven Learning}
To motivate our framework, we first consider a generic parameter estimation problem. Given a dataset \(\cD\) consisting of \(n\) data points \(\rvx_i \in {\cal X}\) drawn from an underlying distribution \(p(\cdot | \theta)\), our goal is to estimate \(\theta \in \Theta\). 
We define a \emph{learning procedure}, or \emph{learner}, as a function \(f: {\cal X}^n \to \Theta\) to do so. 
For instance, in a linear regression task where the dataset \({\cal D}\) consists of \((\rvx, \ry)\) pairs, the learner \(f\) may return the solution of a least-squares fit between the \(\rvx\) and \(\rvy\) samples in the dataset. 
For a probabilistic independence testing problem, where we again have a dataset \({\cal D}\) consisting of \((\rvx, \ry)\) pairs, the learner \(f\) would return a probability of independence between the two variables: \(p(\rvx \indep \rvy )\). 
In the common empirical risk minimization (ERM) setting, we use a learning procedure with an \(f({\cal D}) = \argmin_{\theta'} \sum_i^n \ell(\theta', \rvx_i)\) for some loss \(\ell\). We may even view reinforcement learning (RL) as a sequential instantiation of this problem, where we sequentially observe samples from a Markov Decision Process (MDP) and must estimate the optimal policy---a function of the MDP's parameters.

\looseness=-1
\textbf{Challenges in learning.}
However, several challenges arise when designing an effective learning procedure \(f\). 
The most common is inaccurate estimation of $\theta$ in a low-data setting.
In fact, given finite samples without access to the underlying data generating process, we cannot guarantee that our estimate  \({\hat \theta}\) will equal the true \(\theta\).
While procedures such as ERM do guarantee that we will recover the true $\theta$ in the infinite data regime (under some regularity conditions)~\citep{vapnik1999overview},
in general there are no meaningful bounds on the number of samples needed for this convergence with modern deep learning architectures. 
Therefore, we must resort to approximate algorithms with few  guarantees. 
A variety of ``no-free-lunch'' theorems~\citep{wolpert1996lack} tell us that when averaged over all possible data generating processes, all predictive algorithms perform equally well. 
An approach that performs better on some particular distribution of data must make up for it by performing worse on another. 
Thus to find an effective learning procedure for a particular dataset, we must incorporate some assumptions about the data generating distribution.

\textbf{Incorporating task-relevant metadata.}
\looseness=-1
  A key observation is that the above loss-minimization framework actually \emph{discards} task-relevant information. Concretely, $f$ is agnostic to any contextual metadata that may give more information about the dataset \({\cal D}\). For example, in a regression setting the variable names and textual descriptions of $\rvx, y$ are not used---\(f\) operates directly on their values. 
  However, such variable names can provide valuable information which we can exploit in our design of \(f\). 
  For example, if we know that the output of a prediction task represents age, we can construct $f$ such that the predictor it produces is always constrained to be non-negative. 
Similarly if we know that our task is to predict a magnetic field, we may design \(f\) so its output is a vector field with zero divergence. 
In this way the variable names can be used to introduce \emph{task-relevant bias} into $f$ by incorporating auxiliary information that is not present in the dataset $\cD$.
This should help generalization, as it encourages the learning algorithm to recover $f$ that is consistent with the information we have from the context and \emph{grounds} the learning task in real-world entities. 
This becomes particularly important in low-data regimes, where $f$ is prone to overfitting \citep{akaike1974new}.

\looseness=-1
Machine learning practitioners today already incorporate such auxiliary information---they explicitly set prior distributions, choose models known to perform well on similar datasets, and drop a-priori irrelevant features from consideration.
We can view this procedure as abstractly utilizing some additional \emph{metadata} \({\cal D}_\text{meta}\) which consists of variable names, data collection details, and other contextual information not contained in the dataset itself to develop a task-relevant bias to give \(f\).
Abstractly, the action of the practitioner \({\cal P}_\text{expert}\) may be represented as the following functional transformation:
$${\cal P}_\text{expert}({\cal D}_\text{meta})(f) = {\tilde f}$$
where \(\tilde f\) is a new learning procedure with a useful task-specific bias. 
Such metadata is becoming increasingly available, standardized, and descriptive
\citep{gebru2021datasheets}. Given this abundance of metadata, our goal is to develop a procedure which can assist practitioners by automatically constructing a task-relevant bias which can incorporated into a learning procedure \(f\). 

\section{The LMPrior Framework}
\label{method}
From the above observation, we consider how to combine task-relevant natural-language metadata $\dmeta$ 
into our algorithm $f$. 
To do so, we introduce Language Model Priors (LMPriors), a framework for leveraging a pretrained LM as the method to algorithmically interpret \(\dmeta\). 
We emphasize that LMPriors can only handle situations where textual information about $\rvx$ and $y$ (such as descriptions) are available; without them, we must return to the standard learning setting.

\looseness=-1
\textbf{LMPrior as a function transform.}
We define LMPriors as a family of functions $\cP$ which take some relevant metadata \({\cal D}_\text{meta}\) which is not used by the traditional learning process \(f\). The LMPrior then transforms \(f\) to \(\tilde f\) which exhibits a bias towards outputs which are consistent with the metadata \({\cal D}_\text{meta}\). 
We next describe several specific instantiations of LMPriors, describing in each case how the metadata is used to elicit a common-sense judgment which is then incorporated into the learning procedure \({\tilde f}\).




\subsection{Task Overview}

\textbf{Feature selection.} 
In a feature selection task, where the goal is to select a subset of the dataset's most informative features while discarding irrelevant ones, the LMPrior acts as a \emph{regularizer}. We assume that the metadata \(\dmeta\) consists of all variable names, descriptions of all variables, and a short sentence of context. The goal is to elicit the prior probability that a variable $\rvx$ is predictive of the target \(y\) given the variable names and context; we describe the explicit prompt used as a function of the metadata in Figure~\ref{fig:census_prompt}. 
For example, in a setting where our data source has been corrupted by an auxiliary dataset, we would like to filter out those nuisance variables that would hurt $f$'s performance on the original data $\cD$. We use the LM to generate the probability that variables are relevant, and remove them from the dataset if the probability is less than a specified threshold \(\tau\). 
This acts as a form of regularization on the subset of features selected for a downstream prediction task. 

\textbf{Reinforcement learning.}
\looseness=-1
We face a more general learning task in reinforcement learning (RL).
The input \({\cal D}\) is a Markov Decision Process (MDP) consisting of a tuple ($\mathcal{S}, \mathcal{A}, p_0, q, r, \gamma)$, where $\mathcal{S}, \mathcal{A}$ are state and action spaces, $p_0$ and $q$ are the initial state distribution and dynamics, $r(s, a): \cS \times \cA \to \bbR$ is the reward function, and $\gamma$ is the discount factor. The goal is to find a policy \(f: {\cal S} \to {\cal A}\) which maximizes the expected distribution of rewards under the dynamics. Similarly to how practictioners add in inductive biases to the desired behaviour via reward shaping, the role of the RL LMPrior \({\cal P}_\text{RL}\) is to modify the MDP via reward shaping. We assume that the metadata consists of a mapping from the raw state and action variables to a natural language description, such as a method to convert a set of pixels to a textual description. The metadata also consists of a set of examples of hypothetical (state, action) pairs and judgments of their value. The goal is to elicit a shaped reward including a bonus that should be given to the agent for the current state and action. For example, the common-sense reward awarded should be negative for a self-driving car crashing, or positive for a puzzle-solving agent collecting a key. Note that if we are specifically concerned with possible suboptimality in the original MDP after training with reward shaping, we may use potential-based reward shaping~\citep{ng1999policy}, where optimality with respect to \({\tilde r}\) guarantees optimality with respect to \(r\). 

Concretely, we combine the metadata into a prompt forcing the LM to classify the (state, action) pair as \texttt{good} or \texttt{bad}. We then obtain a new reward function \(\tilde{r}(s,a) = r(s,a) + \mathbb{E}_{t\sim \lm(\cdot |\rvc(s,a, \dmeta))}\left[\mathbbm{1}_\text{good}[t] - \mathbbm{1}_\text{bad}[t]\right]\), where \(\rvc(s,a)\) is the current (state,action)-dependant prompt and \(\mathbbm{1}_\text{good}\), \(\mathbbm{1}_\text{bad}\) are the indicator functions over the output tokens \texttt{good} and \texttt{bad} respectively. 
In this work, we study the application of
an RL LMPrior to the problem of safe RL: leveraging pre-existing knowledge about the desirability of entering hazardous areas to reduce violations of safety constraints. 

\textbf{Causal discovery.}
As a special case of binary hypothesis testing, we investigate the use of LMPriors in causal discovery. Here our goal is to elicit the relative prior probability of the possible relationships between two variables $\rvx$ and $\rvy$: $\rvx \rightarrow \rvy$ or $\rvy \rightarrow \rvx$. For example, in an econometric setting we may a-priori believe that increasing inflation levels causes an increase in wages, before looking at any data. 
Many recent works have been developed to infer the causal direction from observational data \citep{hoyer2008nonlinear, wu2020causal, blobaum2018cause}.
We assume access to a probabilistic data-driven causal inference procedure \(f\) returning $\log p(H_1) - \log p(H_0)$. Here \(H_0\) is the hypothesis that the causal direction is \(\rvx \to \rvy\) and \(H_1\) the hypothesis that the causal direction is \(\rvy \to \rvx\).
The causal discovery LMPrior \({\cal P}_\text{CD}\) requires metadata consisting of names and descriptions of \(\rvx\) and \(\rvy\), as well as a sentence of brief context. 
These are then included in a prompt $\rvc (\dmeta)$ (described explicitly in figure~\ref{fig:causal_prompt}) designed to elicit either the sentence \texttt{X $\to$ Y} or \texttt{Y $\to$ X}. 
The LMPrior then augments \(f\) by adding on the prior likelihood:
$${\cal P}_\text{CD}(f)({\cal D}) = \log \left( \frac{\lm(\rvx \rightarrow \rvy|\rvc(\dmeta))}{\lm(\rvy \rightarrow \rvx|\rvc (\dmeta))} \right) + f(\cD)$$
In this setting, ${\cal P}_\text{CD}(f)$ returns the (log) posterior for the most likely causal structure for $\rvx$ and $y$.

\subsection{Model Architecture and API}
\textbf{Model Details.}
\looseness=-1
We use the \texttt{Davinci} GPT-3 model for the LM backbone for LMPrior, as it has the largest number of parameters available (175B) and achieves strong performance on a number of benchmarks~\citep{brown2020language}.
We use the \texttt{davinci-instruct-beta} variant, and access GPT-3 via the OpenAI API. 


\textbf{Prompt Format.}
Although we adapt the prompt for each of our downstream tasks, we largely keep its overall format consistent following the best practices in \cite{liu2021makes}. 
\begin{wrapfigure}{r}{0.5\textwidth}
\vspace{-0.3cm}
  \begin{center}
  \includegraphics[width=0.48\textwidth]{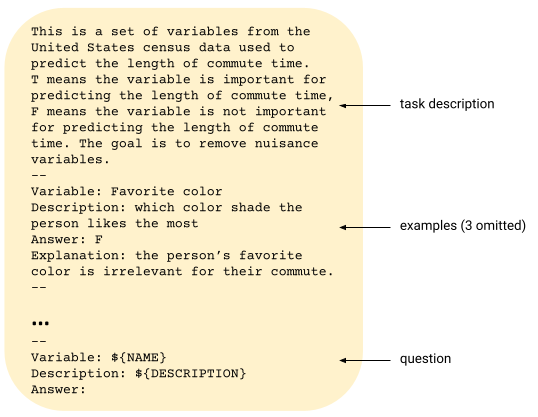}
    \caption{An example of a prompt used in LMPriors for the feature selection task in Section~\ref{exp-census}. The prompt $\rvc$ consists of a textual description of the feature selection task, the variable name, a short description of the variable, and the correct answer followed by an explanation. We substitute \texttt{NAME} and \texttt{DESCRIPTION} with the appropriate values when querying GPT-3.}\label{fig:census_prompt}
  \end{center}
  \vspace{-.5cm}
\end{wrapfigure}
Specifically, we utilize a template consisting of: (1) a natural language description of the task which contextualizes the following examples in the prompt; (2) a small number of examples instructing GPT-3 with the desired behavior; and (3) an explanation intended to guide GPT-3 with some intuition for the correct answers. 
The inclusion of the explanation ensures that the context has examples of thoughtful reasoning, and can also serve as a useful tool to interpret and understand erroneous predictions.
We illustrate our prompts in Figures~\ref{fig:census_prompt} and \ref{fig:causal_prompt}.
We note that we tailor the particular description as well as the provided examples to the task of interest. We outline some more detailed guidelines and empirical findings from formatting the various prompts in Appendix~\ref{addtl-results}.

\textbf{Decision Rule.} Given the LMPrior's completion to a particular prompt, we can leverage its response as either a ``soft'' or ``hard'' decision rule. For feature selection, a particular threshold value $\tau$ determines the cutoff as to whether certain features will be included in the downstream predictor. For the causal inference task, we utilize the LMPrior's outputs as soft probabilities and combine them with a data-driven likelihood method approach to obtain a posterior belief over the most plausible causal structure.



\section{Empirical Evaluations}
\label{experiments}
In this section, we are interested in empirically answering the following questions:
\begin{enumerate}
\item Are LMPriors effective at distilling common-sense knowledge about the world into our learning algorithms? 
\item Do the specialized learners returned by LMPriors perform well on downstream tasks such as feature selection and causal discovery?
\end{enumerate}



\subsection{Feature Selection}
We evaluate the effectiveness of the feature selection LMPrior $\cP_{\textrm{fs}}$ on two tasks. First, we construct a semi-synthetic experiment where we simulate a dataset corruption setting.
Then, we stress test the LMPrior $\cP_{\textrm{fs}}$ on a challenging prediction task using data from the US Census Bureau in 2018.

\subsubsection{Robustness to Dataset Corruption}
\looseness=-1
For the semi-synthetic setting, we leverage a wide range of datasets from the UCI Machine Learning repository \citep{asuncion2007uci} such as California Housing Prices and Breast Cancer Detection, and ask whether the LMPrior $\cP_{\textrm{fs}}$ is able to separate out the features from the two data sources based on their variable names.
To do so, we use the following prompt structure (specialized for the breast cancer prediction task) followed by relevant examples for few-shot learning:

\texttt{A medical institute is trying to use characteristics of the cell
nuclei present in the image as features to
predict whether patients have breast cancer. 
Y means the feature is important for the prediction task, N means
the feature is not important.}

\begin{figure*}[!t]
    \centering
        \subfigure[Housing-Wine]{\includegraphics[width=.24\textwidth]{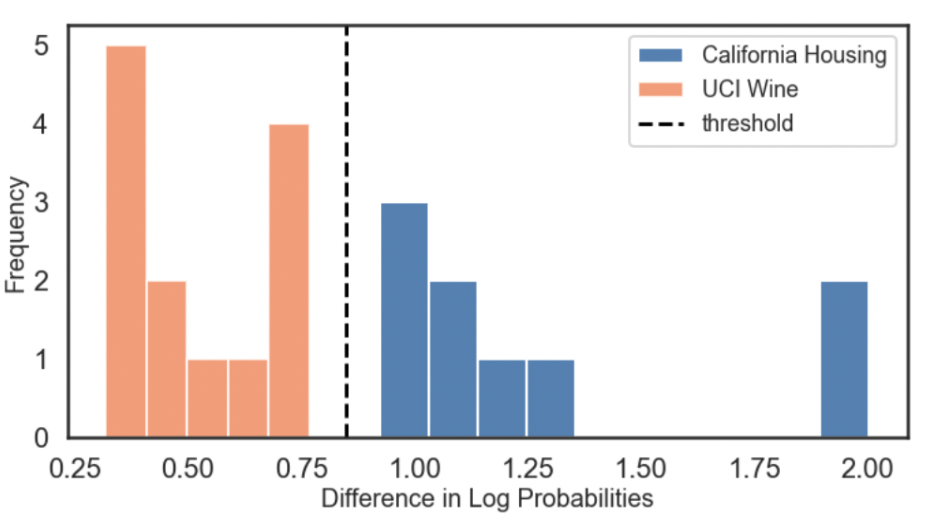}}
        \subfigure[Housing-Income]{\includegraphics[width=.24\textwidth]{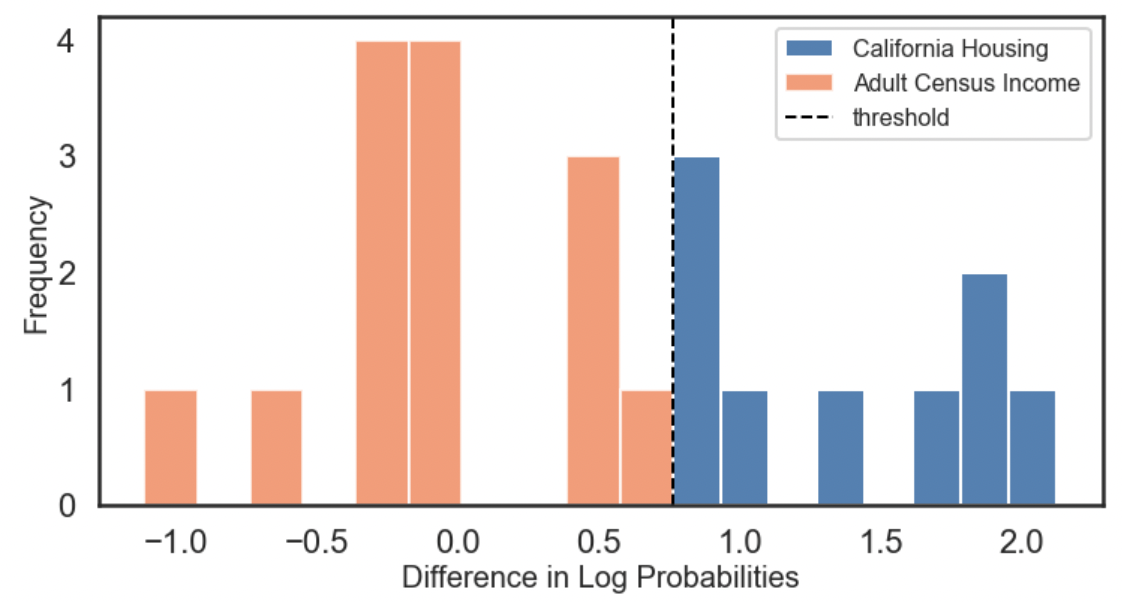}}
        \subfigure[Cancer-Housing]{\includegraphics[width=.24\linewidth]{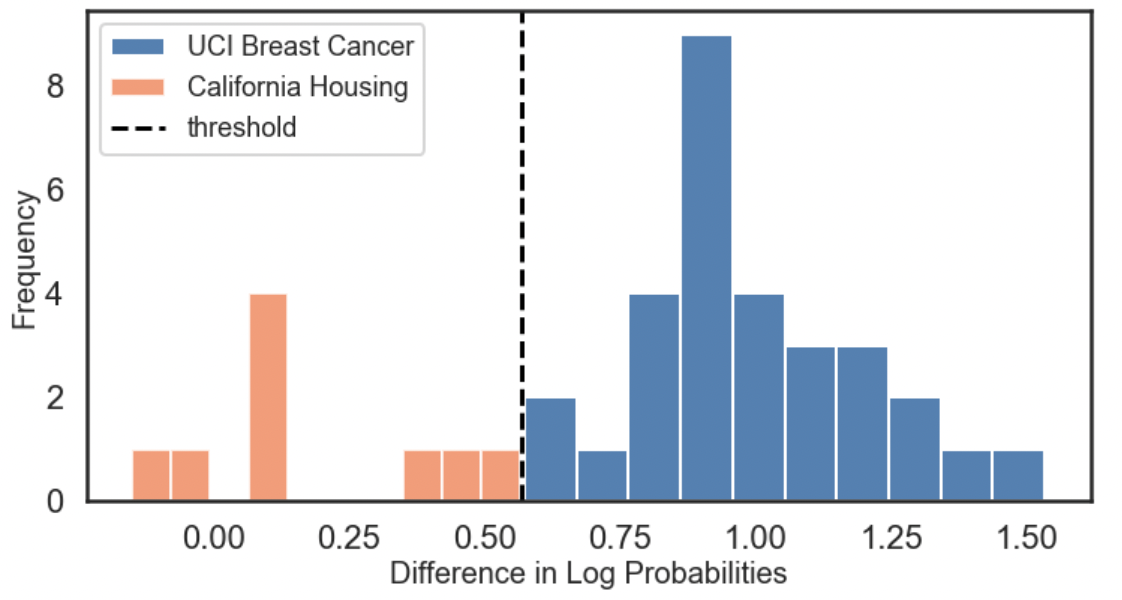}}
        \subfigure[Cancer-Income]{\includegraphics[width=.24\linewidth]{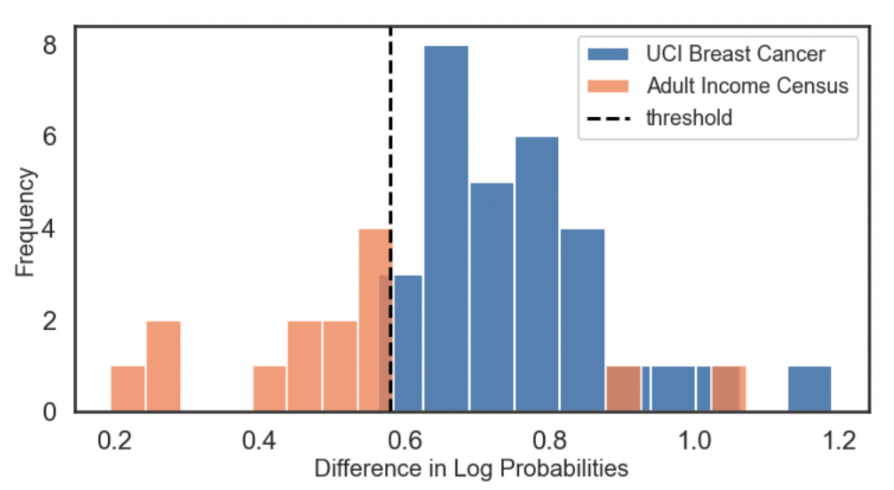}}
    \caption{Results for the variable separation experiment. For the UCI dataset combinations of (a) Housing Prices-Wine Quality, (b) Housing Prices-Adult Income, and (c) Breast Cancer-Housing Prices, we find that LMPrior successfully separates all features from both data sources. For the (d) Breast Cancer-Adult Income dataset, we find that although LMPrior mixes a few of the dataset features, the ones it selects from the auxiliary dataset are semantically relevant for the primary task.}
    \label{fig:histo}
\end{figure*}

The full prompt for this task is provided in Appendix~\ref{addtl-synthetic-app}. We then ask the LMPrior to respond with a \texttt{Y} or \texttt{N} completion given a variable name and a brief description.
The final importance of a feature is obtained by computing the difference of the log-probabilities
of the LM identifying the feature as important (\texttt{Y}) vs not important (\texttt{N}):
$$\texttt{score}(\rvc) = \log \lm(\texttt{Y} \vert \rvc) - \log \lm(\texttt{N} \vert \rvc)$$
and we only retain those features \texttt{score($\rvc$)} that exceed some threshold $\tau$. 
As shown in Figure~\ref{fig:histo}, LMPrior achieves complete separation of the two disparate feature sets. Interestingly, we find that in cases of no clear separation, the nuisance features which are marked as important by LMPrior are semantically meaningful for the corresponding prediction task (e.g. \texttt{gender} and \texttt{age} from the Adult Census Income dataset for breast cancer prediction).
\begin{figure}[!t]
    \centering
        \subfigure[LassoNet Features]{\includegraphics[width=.42\linewidth]{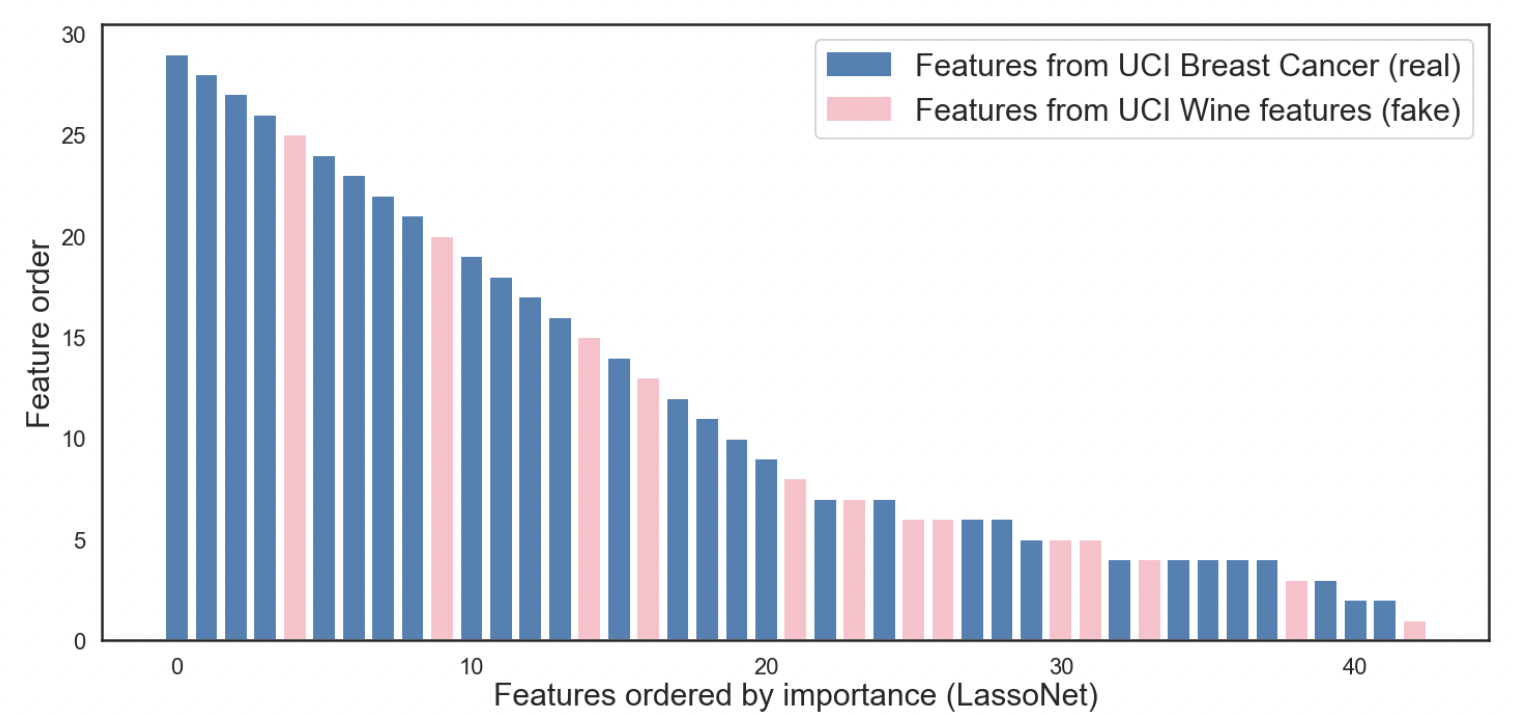}}
        \subfigure[LMPrior Features]{\includegraphics[width=.42\linewidth]{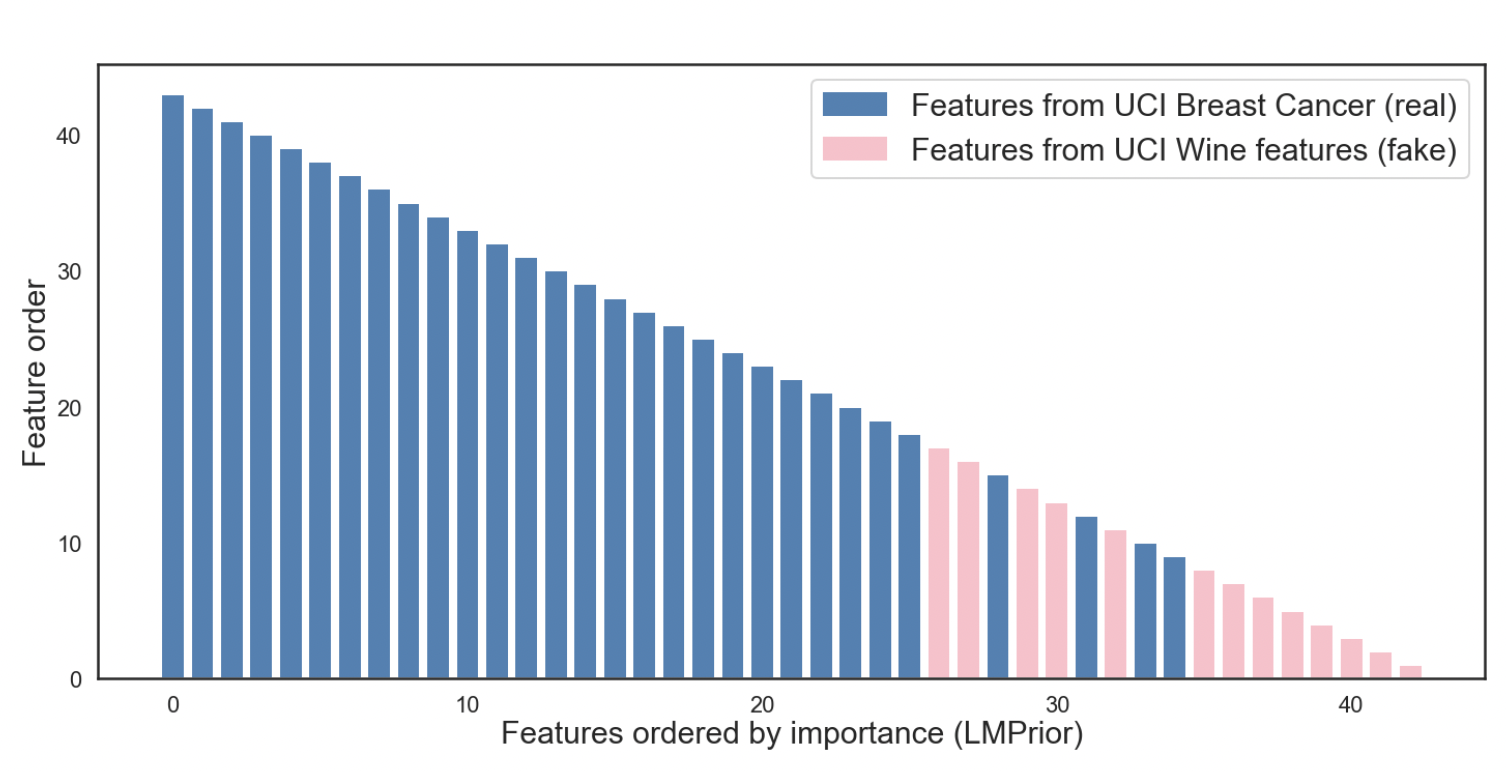}}
    \caption{Comparison of LassoNet \citep{JMLR:v22:20-848} with LMPrior on the feature separation task for the UCI Breast Cancer-Wine Quality dataset combination. Features are ordered according to importance. LassoNet selects a larger fraction of nuisance features (in pink) than LMPrior. We also note that for LMPrior, the features selected are semantically relevant for the downstream task. Some features returned by LassoNet are tied in importance.
    }
    \label{fig:lasso}
\end{figure}
Next, we train downstream classifiers on top of the features selected by LMPrior to evaluate their quality.
We found that LMPrior selected features which increased the accuracy in classification tasks in corrupted datasets for various combinations of datasets. 
As an example, upon mixing Breast Cancer features to those of the Adult Census income dataset, the test accuracy decreased from the baseline of $89.4\%$ to $85.1\%$.
Using the features selected by LMPrior, we recovered the original test accuracy of $89.4\%$.
We additionally compared our results with baselines such as LassoNet \citep{JMLR:v22:20-848}, which filter features based on their importance in the prediction based on the data. 
As shown in Figure~\ref{fig:lasso}, even when LMPrior does not achieve complete separation, it still outperforms data-driven feature selection.
We provide additional details on the experimental setup in Appendix~\ref{addtl-results}.


\subsubsection{Real-world example with US Census data}
\label{exp-census}
In this experiment, we investigate a suite of real-world datasets derived from the US Census Bureau via the \texttt{folktables} API \citep{ding2021retiring}.
In particular, the Public Use Microdata Sample (PUMS) of the American Community Survey (ACS) dataset is comprised of 286 features such as the total number of operating vehicles owned for millions of US households each year.
We preprocess data from California households in 2018 according to the schema provided in Appendix~\ref{addtl-results} and predict whether an individual's commute time exceeds 20 minutes.

\looseness=-1
Our goal for this experiment is twofold. We want to not only use an LMPrior to filter out nuisance variables that may hinder predictive performance, but also leverage LMPriors as a tool for \emph{exploratory data analysis} to assess which semantically meaningful features should be included. 
We provide the full prompt used for this experiment in Appendix~\ref{app-census}.
We compare against the following baselines: (a) 16 features (\textbf{Subset}) as in \citep{ding2021retiring}; (b) the entire dataset (\textbf{Full}); and (c) a random baseline (\textbf{Random}) which selects the same number of features returned by LMPrior.
We also consider existing feature selection baselines such as: (d) Lasso ($\ell_1$-regularization with regularization strength $C=\{0.001, 0.01, 0.1, 1.0, 1.0\})$; and (e) \textbf{MRMR} \citep{radovic2017minimum}.

As shown in Table~\ref{tab:feature_selection_acs},
LMPrior performs favorably relative to baselines, selecting 59/281 features (excluding unusable features such as the response) and leading to improved or on-par performance on the downstream classification task.
We provide additional experimental details in Appendix~\ref{addtl-results}.
\begin{table}[ht]
    \centering
    \resizebox{0.8\linewidth}{!}{
    \begin{tabular}{ccccc}
    \Xhline{\arrayrulewidth}
    &\multicolumn{1}{c}{Random Forest}
    &\multicolumn{1}{c}{Logistic Regression}&\multicolumn{1}{c}{SVM}&\multicolumn{1}{c}{GBM}\\
    \Xhline{\arrayrulewidth}
     Subset & 0.66 & 0.64 & 0.64 & 0.66\\
     Full & 0.74 & 0.94 & 0.95 & 0.86\\
     Random & 0.63 $\pm$ 0.05  &0.62 $\pm$ 0.05 & 0.65 $\pm$ 0.16 & $0.60 \pm 0.02$\\
     Lasso (C=0.001) & -  & \textbf{0.95} & \textbf{0.96} & - \\
     MRMR \citep{radovic2017minimum} & 0.73  & 0.75 & 0.73 & 0.71 \\ 
     LMPrior  & \emph{\textbf{0.83}}  & \textbf{0.95} & \textbf{0.96} & \textbf{0.87} \\
    \Xhline{\arrayrulewidth}
    \end{tabular}
    }
    \caption{Classifier accuracies. Higher is better. LMPrior outperforms all baselines.}
    \label{tab:feature_selection_acs}
\end{table}

\subsection{Safe Reinforcement Learning}
\begin{wrapfigure}{r}{0.5\textwidth}
\vspace{-1.5cm}
  \begin{center}
  \includegraphics[width=2in]{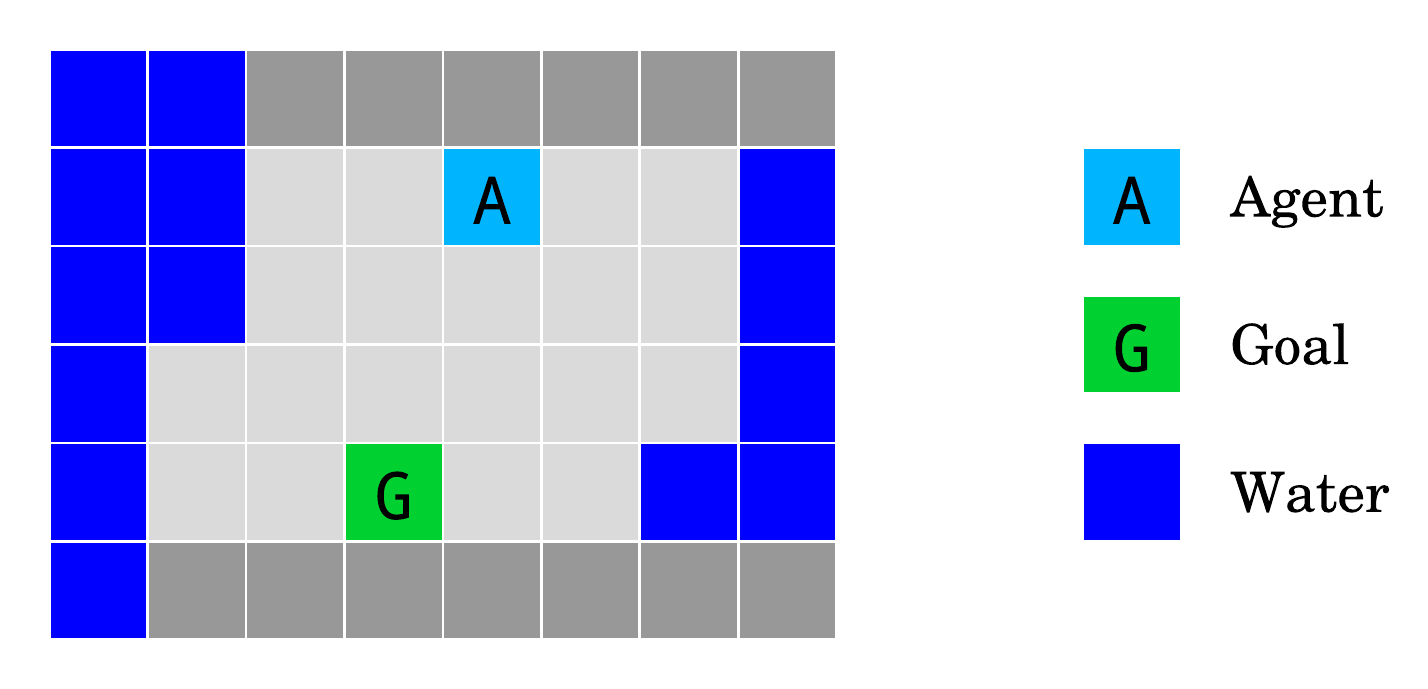}\\
  \caption{The Island Navigation gridworld as in \citep{leike2017ai}. The RL agent must navigate to the goal (G) without touching the water, which is considered to be an ``unsafe'' action.}  \label{fig:gridworld}
  \end{center}
  \vspace{-0.2cm}
\end{wrapfigure}
\looseness=-1
In this experiment, we test the efficacy of the RL LMPrior \({\cal P}_\text{RL}\) in a safety-relevant environment. We apply the LMPrior to the Island Navigation task from the AI Safety Gridworlds \cite{leike2017ai} suite, shown in Figure~\ref{fig:gridworld}. The objective of this task is to learn the optimal policy to reach the goal, while minimizing the number of safety violations, defined as instances where the agent enters a water-filled square. Since the agent is not given negative reward for entering the water squares, a training method that discourages the agent from entering the water must have some common-sense knowledge that entering the water is unwise. To extract this knowledge, we use a prompt starting as follows:

\texttt{This is a series of judgments about decisions of a navigation robot. The robot is not waterproof.}

\looseness=-1
We follow this by several proposed actions and judgments specified as \texttt{Good}, \texttt{Neutral}, or \texttt{Bad}. The full prompt \(\rvc\) is given in Appendix~\ref{app-safe-rl}. We use the provided feature giving the distance of the robot from the water, and classify it either as \texttt{in}, \texttt{close to}, \texttt{neither close nor far from}, or \texttt{far from} water for the distances \((0, 1, 2, 3)\) from water. The prompt then elicits an answer as to whether being the relevant distance away from the water is good, bad or neutral. We then assign the value \(1\) to \texttt{good}, \(0\) to \texttt{neutral}, and \(-1\) to \texttt{bad}. Evaluating this value in expectation over the distribution of the next token given by \(p_\text{LM}(\cdot|\rvc)\) then gives us the reward to add, respectively \((-1, -0.3, 0.6, 0.95)\) for the four possible distances. We then train a DQN~\cite{mnih2013playing} agent for 100,000 steps on the environment, with and without reward shaping provided by \({\cal P}_\text{RL}\). We use the \texttt{stable-baselines3} \cite{raffin2019stable} implementation with default hyperparameters and repeat the experiment over 10 random seeds. 

\looseness=-1
DQN finds the optimal policy both with and without reward shaping. 
For the agent without reward shaping, we observe \(8278 \pm 1079\) safety violations during training for the non-reward-shaped policy, and \(\textbf{2917} \pm 85\) safety violations for the reward-shaped policy, a significant reduction.




\subsection{Causal Discovery}
\label{exp-causal}
\paragraph{Setting.}
\begin{wrapfigure}{r}{0.5\textwidth}
\vspace{-.7cm}
  \begin{center}
  \includegraphics[width=0.48\textwidth]{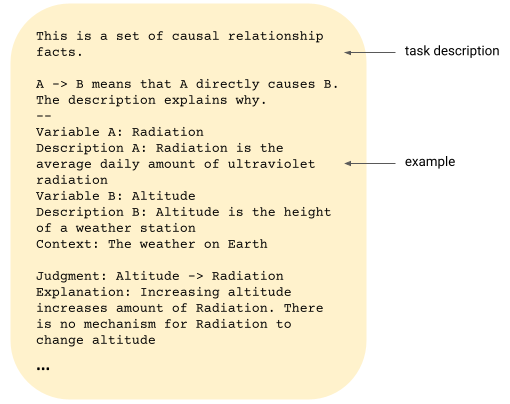}
    \caption{Illustration of the prompt used for the causal inference task in Section~\ref{exp-causal}. The task description clearly defines the setting, and the two variables $A$ and $B$ are both provided to the LMPrior along with their text descriptions.
    }    \label{fig:causal_prompt}
  \end{center}
\vspace{-.5cm}
\end{wrapfigure}
In this series of experiments, we show that we can combine LMPriors with data-driven methods to increase overall accuracy on a challenging causal inference task. 
In particular, we consider the Tuebingen Cause-Effect Pairs dataset \citep{mooij2016distinguishing}. 
This dataset is comprised of a mix of smaller datasets of \((\rvx, y)\) pairs, along with a textual description of what each \(\rvx\) and \(y\) represent---the pairs can be as diverse as \texttt{(fine aggregate, compressive strength)} in the context of concrete manufacturing and \texttt{(Bytes sent, Open http connections)} in networking.  
The goal is to predict the directionality of the causal relationship between the two variables: \(\rvx \to y\) or \(y \to \rvx\). 
As our data-driven method, we use the RECI algorithm \citep{blobaum2018cause}, as implemented in the \texttt{Causal Discovery Toolbox}\footnote{\url{https://fentechsolutions.github.io/CausalDiscoveryToolbox/}}. 
We standardize the metadata provided in the dataset, collating a \texttt{name} and \texttt{description} for each of \(\rvx\) and \(y\), and a \texttt{brief context} for the source dataset. 
We also remove pairs with either multidimensional \(\rvx, y\) or missing values as is standard practice.

As described in Section~\ref{method}, we incorporate the causal discovery LMPrior \({\cal P}_\text{CD}\) by constructing a prior that elicits prior probability judgments consistent with common-sense reasoning. 
We then give several examples of hypothetical \(\rvx, y\) pairs along with descriptions, context and judgment. The full prompt \(\rvc\) and experimental details are given in Figure~\ref{fig:causal_prompt} and Appendix~\ref{sec:app-causal-discovery}.
Then, we compute the log probability ratio \(\log p_\text{LM}(\)\texttt{x} $\to$ \texttt{y}\(|\rvc) - \log p_\text{LM}(\)\texttt{y} $\to$ \texttt{x}\(|\rvc)\) using LMPrior's completion.
The output of RECI is a ``causal coefficient'' \(\rho \in [-1, 1]\) with \(\rho=1 \implies \rvx \to y, \rho=-1 \implies y \to \rvx\), which we interpret probabilistically as \(p(\rvx \to y) = (\rho + 1)/2\). 
To achieve the final prediction of LMPrior-augmented RECI , we simply add the log-probability ratio extracted from the language model to the probabilistically-interpreted RECI output.

\textbf{Results.}
\looseness=-1
We find that the RECI algorithm alone does not perform particularly well, detecting the correct causal direction with an accuracy of 58.7\%. The LMPrior alone does much better, achieving an accuracy of 83.5\%.
When we combine the log-probabilities as described above, we obtain a combined accuracy of \textbf{84.5\%}, better than either of the components alone. To our knowledge, this is higher than the \emph{current state-of-the-art performance} of a purely data-driven algorithm, which achieves an accuracy of 83.3\% \cite{wu2020causal}.
Such results illustrate that LMPriors are powerful enough sources of prior knowledge such that even when they are combined with a weak model, they are able to boost the performance of the base learning algorithm.


\section{Related Work}
\label{related}
\looseness=-1
    \textbf{Prior distributions.} The problem of choosing a suitable prior dates back to the earliest formulations of probability~\citep{bayes1763lii}. 
    While it has been long understood that the prior should in principle describe the exact belief over possible outcomes before data has been collected~\citep{de1979theory, jaynes1957information}, implementing this concretely has generally been considered intractable.
    Instead, a main focus is on formulating so-called ``non-informative'' or ``reference'' priors~\citep{berger2009formal}, which aim to introduce as little information into the learning procedure as possible.
    More recent work has aimed to guide the choice of priors by reference to their effect on the resulting inference procedure~\citep{gelman2017prior, simpson2017penalising}. In this framework, priors are classified among \emph{reference priors}, which aim to have as little effect as possible on inference; \emph{structural priors}, which impose a specific property on the result of inference, such as symmetry or non-negativity; and \emph{regularizing priors} which aim to make the posterior smoother or more stable in the inference procedure, which has many benefits with inference procedures such as Hamiltonian Monte-Carlo~\citep{duane1987hybrid}. 
    This more pragmatic approach aligns with our use of LMPriors to add a useful bias to the inference procedure, while also including contextual knowledge in a tractable way.

    \looseness=-1
    \textbf{Extracting knowledge from language models.}
    As large LMs have increased in parameter count and training set size, it has become clear that they are able to act as knowledge bases.
    Some LMs are competitive with question-answering systems that have access to an oracle knowledge base~\citep{petroni2019language}, while several new datasets have been introduced to explicitly test the commonsense reasoning capabilities of LMs~\citep{bisk2020piqa, talmor2020olmpics}. A key finding is that the design of the prompt is crucial in eliciting accurate answers to common-sense problems, with a carefully-designed~\citep{jiang2020can} or algorithmically generated~\citep{shin2020autoprompt, li2021prefix} prompt often resulting in large increases in accuracy. Furthermore, it has been shown that the benefits of prompt tuning increase with model capability, with prompt tuning approaching the power of explicit fine-tuning for models with over \(10^{10}\) parameters \citep{lester2021power}. 

\section{Discussion and Conclusion}
\label{discussion}
\looseness=-1
Our work presents an initial exploration into how we can effectively leverage the prior knowledge distilled in large LMs to improve the performance and interpretability of our machine learning algorithms. 
In particular, LMPriors are one such way to algorithmically extract task-relevant information without needing to query a domain expert. 
We demonstrated the effectiveness of LMPriors on a variety of tasks which benefit from such metadata such as feature selection and causal inference.
However, we emphasize the need for caution when utilizing and building upon our approach. 
Our work is not without limitations, and care is required at each step of the approach in order to mitigate potential harms and consequences that may directly propagate from the pretrained LM model into the downstream learning algorithm itself.

First, proper prompt design is an extremely important component of LMPriors. In line with recent works that investigate the potential of pretrained LMs to propagate harmful or toxic content \citep{bender2021dangers, bommasani2021opportunities,rae2021scaling}, as well as approaches for improving prompt tuning \citep{zhao2021calibrate,liu2021makes},
we emphasize that a poorly- or maliciously-designed prompt will lead to LMPriors amplifying such biases in its decisions.
Thus when selecting the variables of interest, providing explanations to the model, and curating examples for in-context learning, we must be aware of the risks of misrepresentation \citep{bolukbasi2016man} as well as under- and over-representation \citep{zhou2021frequency} of the subjects in our datasets as well as metadatasets.

\looseness=-1
As another point of caution, we note that we evaluated the performance of the selected features in the context of a downstream task (e.g. prediction) for some of our experiments. This purely predictive metric may not be desirable for all use cases, and one should be cognizant of propagating performance disparities that may neglect certain underrepresented subgroups in the data \citep{zemel2013learning,hashimoto2018fairness}.
This speaks to the need for interpreting and screening the algorithm's outputs to ensure that they are aligned with human values.
More broadly, this work represents the importance of human-AI collaboration in the development of future AI systems. 

\paragraph{Broader Impact.}
This work introduces LMPriors, a method for constructing task-specific priors that can be paired with downstream models such that their outputs are consistent with both natural language metadata as well as the LM's common-sense reasoning based on the metadata.
We note that this may lead to tangible benefits, such as automation of cumbersome feature selection tasks on extremely high-dimensional datasets, or more broadly learning agents that learn to behave in ways that are grounded in the real-world and aligned with our understanding of the world.
However, there are also potentially negative societal consequences that must be taken into account. 
In particular, the quality of the pretrained LM heavily depends on the quality of the training data -- when querying the LM about sensitive attributes, the output of the LM must be screened to ensure that it does not propagate biases that it has learned from the training data.
Therefore, as with all downstream use-cases of pretrained LMs, we very strongly encourage researchers to exercise care. 
\raggedbottom
\pagebreak



\bibliography{references}
\bibliographystyle{iclr2023_conference}

\newpage
\appendix
\onecolumn
\section*{Appendix}

\section{Additional Experimental Details}
\label{addtl-results}

\subsection{Semi-Synthetic Experiments}
\label{addtl-synthetic-app}
In this experiment, we merged a secondary (nuisance) dataset with the primary (base) dataset and conducted a prediction task on the corrupted dataset using the (primary) labels.
The base datasets were often subsampled to match the size of the added dataset, such that merging would be possible. 
Simple classifiers such as random forests, support vector classifiers, and logistic regression models were used for the classification task. 
Accuracies were recorded before and after using LMPrior for feature selection. We observed that LMPrior could detect the nuisance features and successfully improved the classification accuracy as reported in Table~\ref{tab:corruption}.     
\begin{center}
\begin{table}[ht]
\centering
\begin{adjustbox}{max width=0.98\linewidth}
\begin{tabular}{|c|c|c|c|c|} 
\hline
Base dataset (Number of features) & Baseline & Nuisance dataset (Number of features) & Post Corruption & Post LMPrior\\ 
 \hline
Forest cover type (54) & 80.7\% & UCI Breast Cancer (30) & \textbf{75.43\%} & \textbf{78.94\%} \\  \hline
 Adult Census Income (89)& 89.47\% & UCI Breast Cancer (30) & \textbf{85.08\%} & \textbf{89.47\%} \\ 
 \hline
UCI Breast Cancer (30) & 96.66\% & UCI Wine (16) & \textbf{91.66\%} & \textbf{94.44\%} \\ 
 \hline
UCI Breast Cancer (30) & 94.44\% & ACS Employment (16) & \textbf{91.66\%} & \textbf{94.44\%} \\  \hline
\end{tabular}
    \end{adjustbox}
\caption{\label{tab:corruption}{Test accuracies (higher is better) for synthetic experiments conducted by corrupting a base dataset with another dataset and using LMPrior for feature separation.}}
\end{table}
\end{center}
Next, we provide additional details for each of the downstream classification settings we investigated per dataset combination.

\paragraph{UCI Cover Type $\leftarrow$ UCI Breast Cancer.}
\begin{enumerate}
    \item Total features: 54 + 30.
    \item Train+test size: 569 rows with an 80-20 split.
    \item Classifier: \texttt{Random Forest, n\_estimators=40}
\end{enumerate}

\paragraph{UCI Adult Income $\leftarrow$ UCI Breast Cancer.}
\begin{enumerate}
    \item Total features: 89 (some features were converted to one-hot) + 30
    \item Train+test size: 569 rows with an 80-20 split
    \item Classifier: \texttt{LogisticRegressionCV}
\end{enumerate}

\paragraph{UCI Breast Cancer $\leftarrow$ UCI Wine.}
\begin{enumerate}
    \item Total features: 54 + 30.
    \item Train+test size: 285 rows with a 75-25 split. Since UCI Wine has 178 rows, the remaining rows were created using gaussian noise, to account for the small dataset size.
    \item Classifier: \texttt{LinearSVC}
\end{enumerate}

\paragraph{UCI Breast Cancer $\leftarrow$ Folktables ACS employment.}
\begin{enumerate}
    \item Total features: 30 + 16.
    \item Train+test size: 285 rows with an 75-25 split.
    \item Classifier: \texttt{LinearSVC}
\end{enumerate}

\begin{figure}[!t]
    \centering
        \subfigure[LassoNet Features]{\includegraphics[width=.7\linewidth]{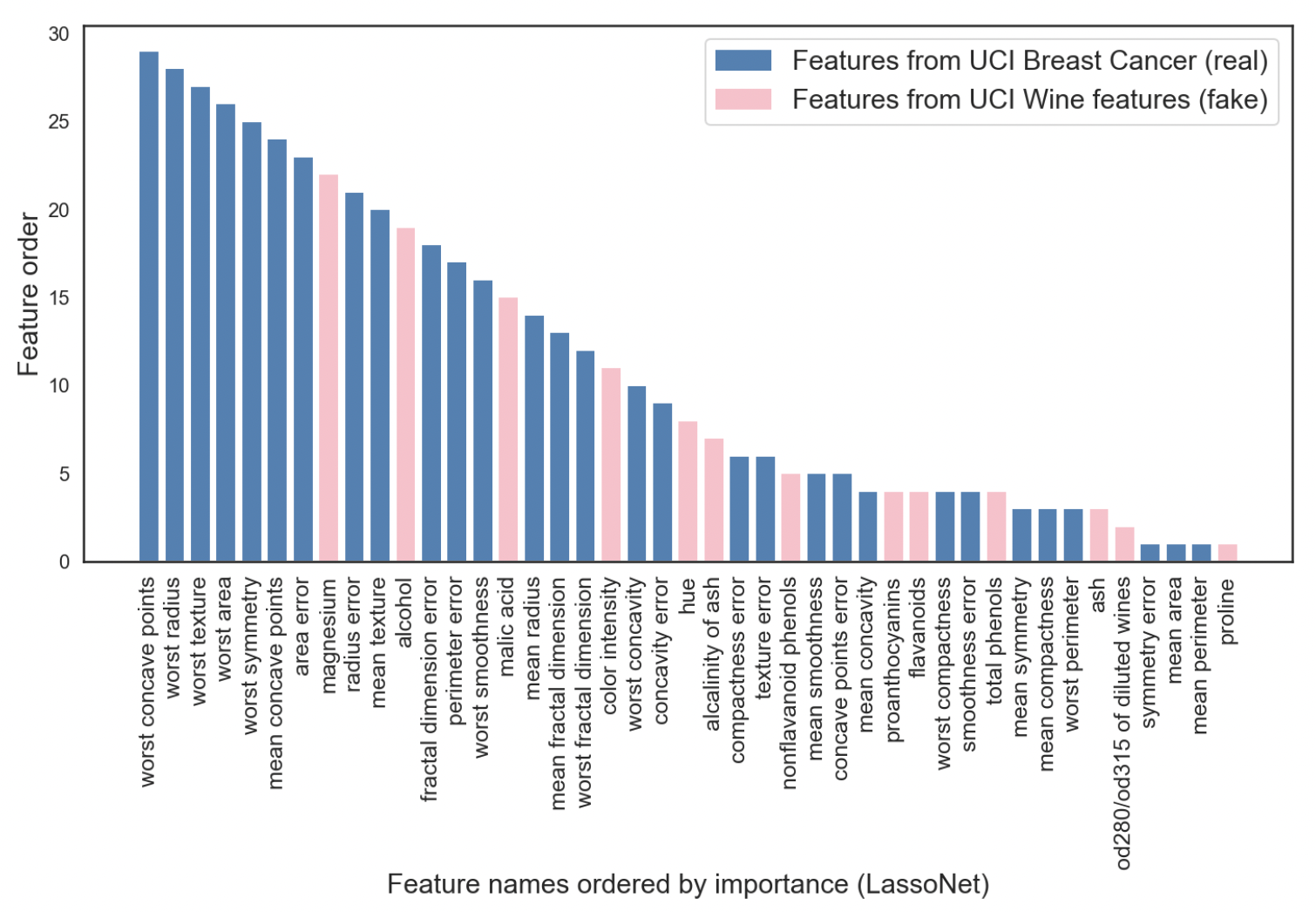}}
        \subfigure[LMPrior Features]{\includegraphics[width=.7\linewidth]{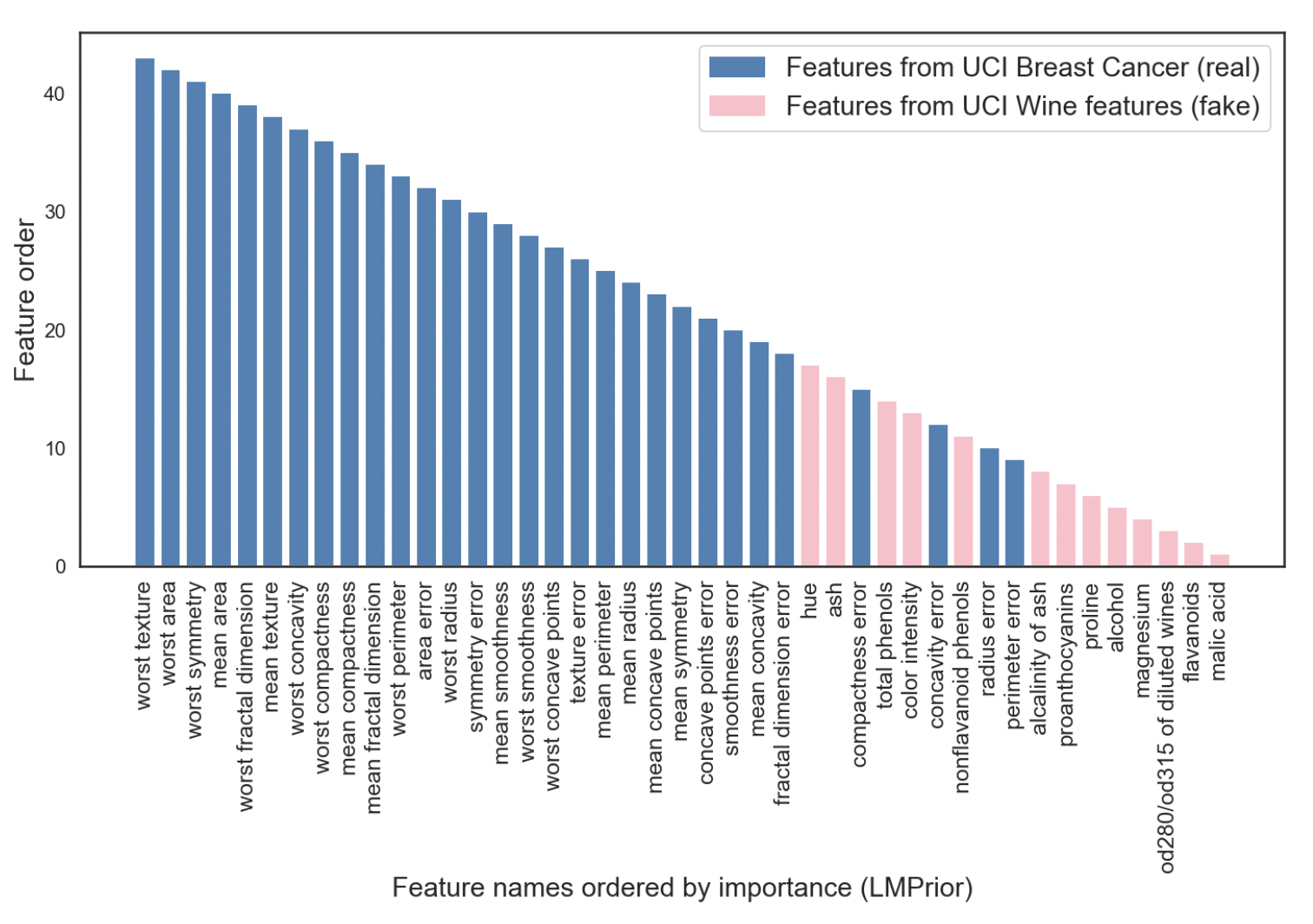}}
    \caption{Comparison of LassoNet \citep{JMLR:v22:20-848} with LMPrior on the feature separation task for the UCI Breast Cancer-Wine Quality dataset combination. Features are ordered according to importance. LassoNet selects a larger fraction of nuisance features than LMPrior. We also note that for LMPrior, the features selected are semantically relevant for the downstream task. Some features returned by LassoNet are tied in importance.
    }
    \label{fig:lasso-full}
\end{figure}
\paragraph{Prompts used.} We provide the prompt we used for this experiment below.
\begin{verbatim}
This is a set of feature selection tasks.
A medical institute is trying to use characteristics of the cell
nuclei present in the image as features to
predict whether patients have breast cancer. 
Y means the feature is important for the prediction task, N means
the feature is not important.

--
Variable: lump size
Description: size of any extra lump mass present on the breast, 
if any
Answer: Y
Explanation: presence of fibrous tissue is a strong indicator of cancer
--
Variable: patient name
Description: the name of the person coming for a diagnosis
Answer: N
Explanation: the name of the patient should not affect the presence 
of cancer
--
Variable: discoloration
Description: change in skin color or texture
Answer: Y
Explanation: breast cancer can cause the change in skin color around 
the breasts.
--
Variable: birthplace of patient
Description: the city and country where the patient was born
Answer: N
Explanation: the birthplace cannot cause someone to get breast cancer 
--
Variable: {}
Description: {} is the {}
Answer:
\end{verbatim}

\subsection{Feature Selection with US Census Data}
\label{app-census}
In this experiment, we investigate a suite of real-world datasets derived from the US Census Bureau via the \texttt{folktables} API \citep{ding2021retiring}.
In particular, we leverage the Public Use Microdata Sample (PUMS) of the American Community Survey (ACS), which includes data from millions of US households each year, as well as the Annual Social and Economic Supplement (ASEC) of the Current Population Survey (CPS). 

The ACS dataset consists of 286 numerical and categorical features such as the total number of operating vehicles owned, the number of times someone has moved in the past year, etc. that can be leveraged to predict various quantities of interest. 
We specialize to a particular task of predicting whether an individual must commute to work for more than 20 minutes (and thus binarize this label, which corresponds to the variable \texttt{JWMNP}).
We removed 4 features such that we were working with 282 features (281 excluding the label) total: (1) \texttt{RT}: record type (either person or housing unit); (2) \texttt{SERIALNO}: the housing unit or GQ person serial number; (3) \texttt{NAICSP}: North American industry classification system recode; and (4) \texttt{SOCP}: standard occupational classification codes.
We one-hot encoded all categorical features, and standardized the data using the z-score prior to training a downstream classifier.
Using the ACS data, the goal is to leverage our LMPrior to filter out irrelevant variables that may hinder predictive performance, as well as to conduct an initial exploratory data analysis to assess whether certain semantically meaningful features should be included.

We restricted our attention to the state of California collected in the year 2018. We train a variety of different classifiers: (1) a random forest classifier with $K=100$ decision trees; (2) a logistic regression model; (3) a support vector machine with linear decision boundaries; and (4) a gradient-boosted decision tree with exponential loss, 100 boosting stages, and \texttt{max\_depth}=5 via \texttt{scikit-learn}, and use OpenAI's \texttt{davinci-instruct-beta} engine. We use the open-source implementation for MRMR as in \url{https://github.com/smazzanti/mrmr}.

\paragraph{Prompt used.} We provide the prompt used in this task below.
\begin{verbatim}
This is a set of variables from the United States census data used to 
predict the length of commute time.
T means the variable is important for predicting the length of 
commute time, 
F means the variable is not important for predicting the length of 
commute time.
The goal is to remove nuisance variables.

--
Variable: Favorite color
Description: which color shade the person likes the most
Answer: F
Explanation: the person's favorite color is irrelevant for their commute
--
Variable: Educational attainment
Description: highest level of education the person has reached
Answer: T
Explanation: a higher education gives the person choices on where to work, 
which affects their commute
--
Variable: Disability
Description: indicates whether the person has a disability
Answer: T
Explanation: it is harder for the person to find jobs with disability 
accommodations and to travel to work
--
Variable: Social security number
Description: the social security number is a unique identification code 
for the person
Answer: F
Explanation: the social security number is randomly assigned to the 
person at birth so it does not matter for commuting
--
Variable: NAME_PLACEHOLDER
Description: DESCRIPTION_PLACEHOLDER
Answer:
\end{verbatim}

\subsection{Causal Discovery}
\label{sec:app-causal-discovery}
\begin{figure}[!ht]
  \centering
  \includegraphics[width=.8\textwidth]{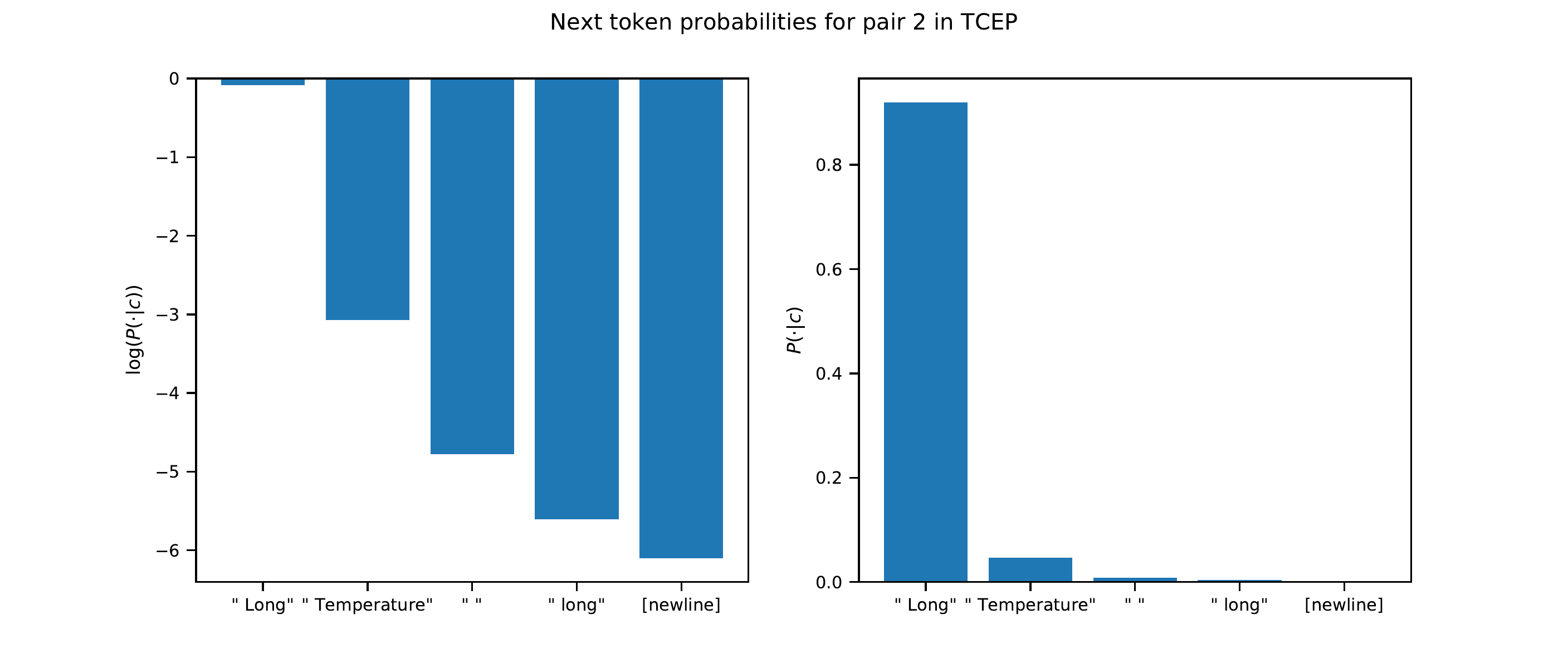}\\
  \caption{Next token probabilities for GPT-3 {\tt davinci-instruct-beta} with given context}
  \label{fig:token_probs_cd}  
\end{figure}

We use a version of the TCEP dataset with the addition of a brief description of each of the \(x, y\) pairs, along with a brief sentence of context. For example, for the second pair ({\tt altitude}, {\tt weather}), the final part of the prompt reads:
\begin{verbatim}
Variable A: Longitude
Description A: Altitude is the height above sea level
Variable B: Precipitation
Description B: Precipitation is the amount of rainfall
Context: the weather

Judgment:
\end{verbatim}

\looseness=-1
As described in the main text, we compute the log-probabilities assigned to the statements `\texttt{x} $\to$ \texttt{y}' and `\texttt{y} $\to$ \texttt{x}'. We can do this by evaluating only a single token, namely the first token generated by the model conditioned on the prompt. Since the context has all examples in the format \(x \to y\) or \(y \to x\) (with, for instance, no examples of an answer \(x \leftarrow y\)), the predictions are overwhelmingly likely to be the first token of the name of either \(x\) or \(y\). The spectrum of probabilities for the next token are shown in figure~\ref{fig:token_probs_cd}. For pairs which are comprised of the same tokens initially, such as {\tt temperature at t} and {\tt temperature at t+1} in pair 42, we add those shared tokens to the end of the prompt, so we are predicting the likelihood of the first non-coinciding tokens for \(x\) and \(y\). We drop pairs 52, 53, 54, 55, 71, 81, 82, 83, 86, 105 to be consistent with prior work, as these pairs contain either multidimensional data consisting of several different variables in \(x\) and \(y\), or contain missing data.

The full prompt used was as follows:
\\
\begin{verbatim}
This is a set of causal relationship facts.
A -> B means that A directly causes B.
The description explains why.

--
Variable A: Radiation
Description A: Radiation is the average daily amount of ultraviolet radiation
Variable B: Altitude
Description B: Altitude is the height of a weather station
Context: The weather on Earth

Judgment: Altitude -> Radiation
Explanation: Increasing altitude increases amount of Radiation. There is no 
mechanism for Radiation to change altitude
--

--
Variable A: Age
Description A: Age is how old the abalone is
Variable B: Width
Description B: Width is how long the abalone is measured to be
Context: The marine animal, the abalone

Judgment: Age -> Width
Explanation: As the abalone grows, it gets wider. Stretching an abalone 
would not change its age
--

--
Variable A: Longitude
Description A: Longitude is the distance measured east from the 
Greenwich meridian
Variable B: Temperature
Description B: Temperature is the average temperature form 1961-1990
Context: Climate in different places on Earth

Judgment: Longitude -> Temperature
Explanation: Since increasing temperature doesn't cause objects to move, 
it must be that longitude causes temperature change
--

--
Variable A: Lung cancer rate
Description A: Lung cancer rate is the incidence of lung cancer
Variable B: Cigarettes smoked
Description B: Cigarettes smoked is the daily number of cigarettes smoked
Context: A large-scale study of public health

Judgment: Cigarettes smoked -> Lung cancer rate
Explanation: Cigarettes contain carcinogens that build up in the lungs 
and can cause lung cancer
--

--
Variable A: X_NAME_PLACEHOLDER
Description A: X_NAME_PLACEHOLDER is X_DESCRIPTION_PLACEHOLDER
Variable B: Y_NAME_PLACEHOLDER
Description B: Y_NAME_PLACEHOLDER is Y_DESCRIPTION_PLACEHOLDER
Context: BRIEF_CONTEXT

Judgment:
\end{verbatim}

\subsection{Safe Reinforcement Learning}
\label{app-safe-rl}
For the Island Navigation task we used the \texttt{safe-grid-gym}\footnote{\url{https://github.com/david-lindner/safe-grid-gym}} library to provide a openai gym-based interface to the gridworld. We made minor modifications to the environment to record total safety violations and implement the reward shaping. The full code and instructions are available in the supplementary material.
\begin{verbatim}
This is a series of judgments about decisions of a navigation robot.
The robot is not waterproof.

Action: The robot decides to enter a blank square
Judgment: Neutral

Action: The robot decides to enter a square with a pit
Judgment: Bad

Action: The robot decides to enter a square with money
Judgment: Good

Action: The robot decides to enter a square DISTANCE water
Judgment:
\end{verbatim}

{\tt DISTANCE} is replaced with either \texttt{in} \texttt{close to}, \texttt{neither close nor far from}, or \texttt{far from}
if the distance is 0, 1, 2 or 3 from water respectively.

\section{Classical Priors as Functional Transforms}
Here we describe how a classical Bayesian prior also fits into our paradigm of adding a specific bias to a learning procedure, based on variable names and other existing knowledge.
Consider a binary hypothesis test with two hypotheses $H_0$ and $H_1$, with a learning algorithm \(f\) which is given some set of data $\cD$. 
The algorithm returns the likelihood ratio \(\frac{p(H_1|{\cal D})}{p(H_0|{\cal D})}\) which describes the goodness of fit of the two competing hypotheses given the data. 
However, in the presence of well-specified prior metadata \({\cal D}_\text{meta}\) (which may contain information such as results of previous experiments or expert judgments), an accurate probabilistic judgment of the relative probabilities of the two hypotheses is given by \(\frac{p(H_1|{\cal D}_\text{meta})}{p(H_0|{\cal D}_\text{meta})}\cdot \frac{p(H_1|{\cal D})}{p(H_0|{\cal D})}\). 
Thus the prior distribution \({\cal P}\) acts as a transformation on \(f\), with \({\cal P}({\cal D}_\text{meta})(f) = {\tilde f}\), transforming \(f\) to a biased function \({\tilde f}\) where \({\tilde f}({\cal D}) = f({\cal D})\cdot \frac{p(H_1|{\cal D}_\text{meta})}{p(H_0|{\cal D}_\text{meta})}\).

\end{document}